\documentclass[conference]{IEEEtran}
\IEEEoverridecommandlockouts
\usepackage{cite}
\usepackage{amsmath,amssymb,amsfonts}
\usepackage{algorithmic}
\usepackage{graphicx}
\usepackage{textcomp}
\usepackage{xcolor}
\usepackage{bm}
\usepackage{booktabs}
\usepackage{multirow}
\usepackage[separate-uncertainty=true]{siunitx}
\usepackage{etoolbox}
\makeatletter
\patchcmd{\@makecaption}
  {\scshape}
  {}
  {}
  {}
\makeatother
\def\BibTeX{{\rm B\kern-.05em{\sc i\kern-.025em b}\kern-.08em
    T\kern-.1667em\lower.7ex\hbox{E}\kern-.125emX}}
    
\ifCLASSOPTIONcompsoc
	\usepackage[caption=false,font=normalsize,labelfont=sf,textfont=sf,subrefformat=parens,labelformat=parens]{subfig}
\else
	\usepackage[caption=false,font=footnotesize,subrefformat=parens,labelformat=parens]{subfig}
\fi

\begin{document}

\title{Federated Learning with Reservoir State Analysis for Time Series Anomaly Detection}
\author{
	\IEEEauthorblockN{
		Keigo Nogami\IEEEauthorrefmark{1}, Hiroto Tamura\IEEEauthorrefmark{2}\IEEEauthorrefmark{3} and Gouhei Tanaka\IEEEauthorrefmark{1}\IEEEauthorrefmark{2}
	}
	\IEEEauthorblockA{
		\IEEEauthorrefmark{1} Department of Computer Science, Nagoya Institute of Technology, Nagoya 466-8555, Japan
	}
	\IEEEauthorblockA{
		\IEEEauthorrefmark{2} International Research Center for Neurointelligence, The University of Tokyo, Tokyo 113-0033, Japan
	}
	\IEEEauthorblockA{
		\IEEEauthorrefmark{3} Graduate Schools for Law and PoliticsThe University of Tokyo, Tokyo 113-0033, Japan
	}
	\IEEEauthorblockA{
		k.nogami.472@stn.nitech.ac.jp, h-tamura@g.ecc.u-tokyo.ac.jp, gtanaka@nitech.ac.jp
	}
}

\maketitle

\begin{abstract}
	With a growing data privacy concern, federated learning has emerged as a promising framework to train machine learning models without sharing locally distributed data.
	In federated learning, local model training by multiple clients and model integration by a server are repeated only through model parameter sharing.
	Most existing federated learning methods assume training deep learning models, which are often computationally demanding.
	To deal with this issue, we propose federated learning methods with reservoir state analysis to seek computational efficiency and data privacy protection simultaneously.
	Specifically, our method relies on Mahalanobis Distance of Reservoir States (MD-RS) method targeting time series anomaly detection, which learns a distribution of reservoir states for normal inputs and detects anomalies based on a deviation from the learned distribution.
	Iterative updating of statistical parameters in the MD-RS enables incremental federated learning (IncFed MD-RS).
	We evaluate the performance of IncFed MD-RS using benchmark datasets for time series anomaly detection.
	The results show that IncFed MD-RS outperforms other federated learning methods with deep learning and reservoir computing models particularly when clients' data are relatively short and heterogeneous.
	We demonstrate that IncFed MD-RS is robust against reduced sample data compared to other methods.
	We also show that the computational cost of IncFed MD-RS can be reduced by subsampling from the reservoir states without performance degradation.
	The proposed method is beneficial especially in anomaly detection applications where computational efficiency, algorithm simplicity, and low communication cost are required.
\end{abstract}

\begin{IEEEkeywords}
	Reservoir Computing, Federated Learning, Time Series Anomaly Detection
\end{IEEEkeywords}

\section{Introduction}

Driven by privacy concerns, a privacy-preserving decentralized approach, called federated learning\cite{FedAvg}, is expected to be a new design for machine learning implementation.
In federated learning, each client trains a deep learning model with its local data, and then a federation of the local models is performed on a central server to aggregate all knowledge among clients.
However, traditional federated learning methods are generally based on deep learning models, which require a large amount of computational resources on the client side.

To address the issue, federated learning with reservoir computing models, such as echo state networks\cite{echo-state-network-jaeger}, is a promising option that can achieve computational efficiency and protect data privacy at the same time.
However, such research is still limited.
For example, Incremental Federated Learning with ESN (IncFed ESN)\cite{IncFed} is the first method that applies federated learning to ESN.
Despite federated learning, constructed models through IncFed ESN are mathematically equivalent to models obtained with the corresponding centralized methods.
Other than such a federated learning method that uses a server with reservoir computing, there are decentralized federated learning methods that work without a server\cite{DIncFed, decentralized-fl-esn}.
Note that all of the methods are designed for time series prediction or classification tasks.
To our best knowledge, no study has applied federated learning with reservoir computing to anomaly detection tasks.

We propose a novel federated learning scheme for time series anomaly detection with reservoir state analyses, called IncFed MD-RS, which bases Mahalanobis Distance of Reservoir States (MD-RS)\cite{MD-RS}.
MD-RS utilizes distributional analysis of reservoir states to detect anomalous data.
Standard anomaly detection methods with reservoir computing are based on a reconstruction approach in which models are trained with normal data to reconstruct input data\cite{schmidl2022anomaly,choi2021deep}.
The anomaly score for unknown input data is given by the reconstruction error, which is expected to be small for normal data due to successful reconstruction and large for anomaly data due to reconstruction failure.
However, the reconstruction often fails for normal data when they are too short or noisy.
Unlike such a reconstruction error-based method, MD-RS uses the distance between the current reservoir state and the distribution of the normal reservoir states for anomaly detection.
It combines the statistical method for anomaly detection using Mahalanobis distance and a reservoir-based nonlinear feature extraction from input data.
During training, MD-RS fits a multivariate Guassian distribution to reservoir state vectors for only normal data in the training phase.
After training, it calculates anomaly scores given by the Mahalanobis distance between a reservoir state vector for test input data and the fitted distribution.
We leverage MD-RS in federated learning settings to perform efficient time series anomaly detection while protecting data privacy.
In particular, Incremental Federated Learning with MD-RS (IncFed MD-RS) enables to construct the same anomaly detection model as that obtained with the centralized method as well as IncFed ESN.

IncFed MD-RS outperforms other federated learning methods for time series anomaly detection, including reservoir-based and deep learning-based methods for every dataset.
Moreover, IncFed MD-RS achieves high performance even with relatively limited training data and is not affected by the number of clients compared to other methods, showing that IncFed MD-RS is well-suited for federated learning, in which many clients collaboratively train a global model.
IncFed MD-RS can also reduce computational costs and communication overhead compared to IncFed ESN without lowering anomaly detection performance.

Our primary contributions are summarized as follows:
\begin{itemize}
	\item We propose the first federated learning scheme with reservoir computing for anomaly detection.
	\item We show that the proposed method can reduce the cost of communications between clients and a server for federated learning compared to the existing method with ESN.
	\item We demonstrate in numerical experiments that the proposed method outperforms other federated learning methods in time series anomaly detection tasks.
\end{itemize}

\section{Background}
\subsection{Echo State Network (ESN)}
ESN is one of the most representative models in reservoir computing and an efficient machine learning method suited for time-series\cite{lukovsevivcius2009reservoir}.
An ESN primarily consists of two layers: the reservoir layer and the readout layer.
The reservoir layer is equivalent to a recurrent neural network with randomly fixed connection weights.
Only the readout parameters are optimized using a regression-based algorithm with a lower computational cost than a gradient-based algorithm.
The reservoir consisting of a sufficiently large number of nonlinear nodes has a role to nonlinearly map inputs into a high-dimensional space and evoke an echo state that preserves the information of past inputs on its own nodes.
Then the readout layer produces an output represented as a weighted sum of the reservoir state vectors.
Unlike conventional recurrent neural network models, the fixed connection weights in the reservoir allow for faster training.
The reservoir state vector $\mathbf{x}(t) \in \mathbb{R}^{N_\mathrm{x}}$ ($N_\mathrm{x}$: the number of reservoir nodes) at time $t$, with random initial condition $\mathbf{x}(0)=\mathbf{0}$, is updated as follows \cite{echo-state-network-jaeger, jaeger2007optimization}:
\begin{equation}
	\begin{split}
		\mathbf{x}(t) & = (1-\alpha) \mathbf{x}(t-1)                                                             \\
		              & \quad + \alpha \tanh (\mathbf{W}_\mathrm{in}\mathbf{u}(t) + \mathbf{W} \mathbf{x}(t-1)),
	\end{split}
	\label{eq:internal-state}
\end{equation}
where $\mathbf{W}_{\mathrm{in}}\in \mathbb{R}^{N_\mathrm{x} \times N_\mathrm{u}}$ ($N_\mathrm{u}$: the number of input nodes) denotes the input-to-reservoir weight matrix, $\mathbf{W} \in \mathbb{R}^{N_\mathrm{x} \times N_\mathrm{x}}$ denotes the reservoir-to-reservoir weight matrix, and $\alpha \in \mathbb{R}$ with $0<\alpha \le 1$ denotes the leaking rate.

The reservoir-to-reservoir weight matrix $\mathbf{W}$ is initialized as a sparse matrix, where only several to 20\% of the elements have non-zero values randomly drawn from a probability distribution such as uniform distribution\cite{lukovsevivcius2009reservoir}.
The reservoir state vector $\mathbf{x}(t)$ is used to determine the output $\mathbf{y}(t) \in \mathbb{R}^{N_\mathrm{y}}$ in the readout layer as follows:
\begin{equation}
	\mathbf{y}(t) = \mathbf{W}_\mathrm{out}\mathbf{x}(t),
	\label{eq:output}
\end{equation}
where $\mathbf{W}_\mathrm{out} \in \mathbb{R}^{N_\mathrm{y} \times N_\mathrm{x}}$ is the output weight matrix.
The matrix $\mathbf{W}_\mathrm{out}$ is normally optimized with a regression-based algorithm using simple matrix operations.
Based on ridge regression with a regularization parameter $\beta$, the optimal weight matrix is calculated as follows:
\begin{equation}
	\mathbf{W}_\mathrm{out} = \mathbf{D}\mathbf{X}^\top(\mathbf{X}\mathbf{X^\top} + \beta \mathbf{I})^{-1},
	\label{eq:ridge-regression}
\end{equation}
where $\mathbf{D} \in \mathbb{R}^{N_\mathrm{y} \times T}$ denotes the target output matrix where target vectors at all time steps are collected column-wise, and $\mathbf{X} \in \mathbb{R}^{N_\mathrm{x} \times T}$ denotes the reservoir state matrix where reservoir states at all time steps are collected column-wise.

\subsection{Federated Learning}
\begin{figure*}[!t]
	\centering
	\subfloat[FedAvg]{\includegraphics[width=0.3\textwidth]{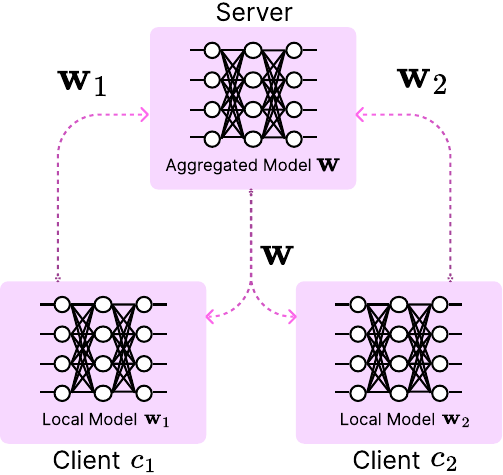}\label{fig:fedavg}}
	\hfil
	\subfloat[IncFed (ESN) ]{\includegraphics[width=0.3\textwidth]{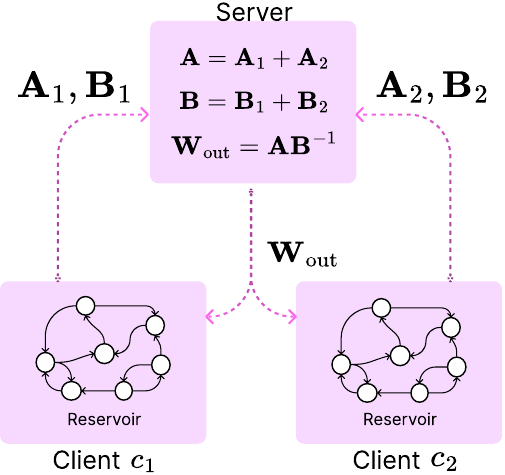}\label{fig:incfed-esn}}
	\hfil
	\subfloat[IncFed (MD-RS)]{\includegraphics[width=0.3\textwidth]{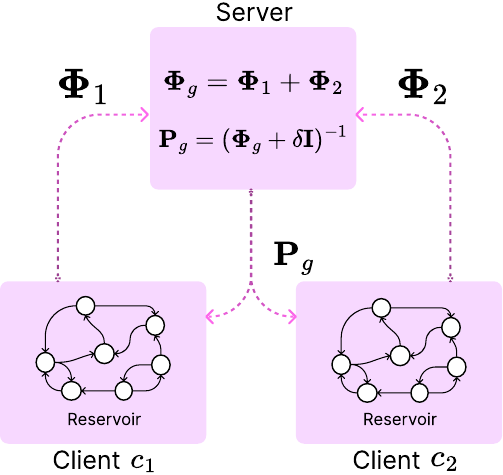}\label{fig:incfed-mdrs}}
	\caption{Schematic illustrations of federated learning schemes.
		(a) Federated Averaging. Each client $c$ sends its local update $\mathbf{w_c}$ to the server. Then the server aggregates all local updates to construct a global model and sends it back to the clients. This procedure continues until training losses converge.
		(b) Incremental Federated Learning with ESN. Each client $c$ sends its matrices containing information on reservoir states and target values. Then the server calculates output weight matrix $\mathbf{W}_\mathrm{out}$ and sends it back to the clients.
		(c) Incremental Federated Learning with MD-RS. Each client $c$ sends its covariance matrix of reservoir state vectors. Then the server calculates the precision matrix by inverse calculation of the aggregated covariance matrix and sends it back to the clients.
	}
	\label{fig:overviews_of_fl_arch}
\end{figure*}

Federated learning is a collaborative approach in which multiple clients train a global model at a server without exchanging their raw data.
By keeping raw data on local devices and aggregating their local models in some way, federated learning helps preserve data privacy and decreases data communication overhead \cite{survey-on-federated-learning, federated-learning-challenges}.
The most representative method in federated learning is Federated Averaging (FedAvg) \cite{FedAvg}.
FedAvg averages the parameters of clients' local models, weighted by the amount of data used for training.
FedAvg proceeds in the following steps \cite{survey-on-federated-learning}:
\begin{enumerate}
	\item The server randomly selects some clients.
	      The most common criteria for client selection lie in being in charge, remaining idle, and on unmetered connections.
	\item The selected clients receive the current global model from the server and initialize their local models with the global model.
	\item Each client trains a local model using its local data.
	      The most common training algorithm for deep neural networks is stochastic gradient descent.
	      To improve the global model, clients send the updated local model to the server.
	\item If most clients successfully complete training, the process continues.
	      However, if many of the clients fail to perform the model training due to factors such as poor network connectivity or limited computational resources, the current round is discarded.
	\item The server aggregates the local models from all the clients through averaging, while considering weighting according to the amount of data provided by each client.
\end{enumerate}
This process is repeated until the training losses converge.
An overview of how the model parameters are transmitted between clients and a server in federated averaging is shown in Fig. \subref*{fig:fedavg}.

FedAvg faces a difficulty when the data distributions of the clients are heterogeneous.
Recent works point out that theoretical guarantee on the convergence of FedAvg does not hold under dataset heterogeneity [12],[13].
To address this issue, FedAvg has been extended to modified versions such as FedProx\cite{FedProx}, SCAFFOLD\cite{SCAFFOLD}, and MOON\cite{Moon}.

\subsection{Incremental Federated Learning with ESN (IncFed ESN)}
IncFed ESN \cite{IncFed} is the first method that trained ESN through federated learning.
It aims to construct a mathematically equivalent model to the corresponding centralized one instead of building a global model by approximating local models as in the FedAvg-like approaches.
This approach is possible because of the two factors in \eqref{eq:ridge-regression}, $\mathbf{D}\mathbf{X}^\top$ and $\mathbf{X}\mathbf{X}^\top$, which can be iteratively updated.
In IncFed ESN, all clients share the same hyperparameters of ESN such as leaking rate $\alpha$, input-to-reservoir weight $\mathbf{W}_\mathrm{in}$, and reservoir-to-reservoir weight $\mathbf{W}$.
Then each client $c$ computes matrices $\mathbf{A}_c \in \mathbb{R}^{N_\mathrm{y} \times N_\mathrm{x}}$ and $\mathbf{B}_c \in \mathbb{R}^{N_\mathrm{x} \times N_\mathrm{x}}$ as follows:
\begin{align}
	\mathbf{A}_c & = \mathbf{D}_c\mathbf{X}^\top_c, \label{eq:Ac}                   \\
	\mathbf{B}_c & = \mathbf{X}_c\mathbf{X}^\top_c + \beta \mathbf{I},\label{eq:Bc}
\end{align}
where $\mathbf{X}_c \in \mathbb{R}^{N_\mathrm{x} \times T}$ is the reservoir state matrix for input training data obtained by client $c$ and $\mathbf{D}_c$ is the target output matrix of client $c$.
After clients send those two matrices to a server, the server sums the matrices from all the clients (denoted by $C$) as follows:
\begin{align}
	\mathbf{A} = \sum_{c \in C}\mathbf{A}_c, \\
	\mathbf{B} = \sum_{c \in C}\mathbf{B}_c.
	\label{eq:A_B}
\end{align}
Then the server computes the readout layer weights in the following equation:
\begin{equation}
	\mathbf{W}_\mathrm{out} = \mathbf{A}\mathbf{B}^{-1}.
	\label{eq:Wout-fed}
\end{equation}

Additionally, it is possible to update the readout layer weights by simply summing them as soon as new data arrive with the associated labels.
We assume that client $c$ has already computed the two matrices $\mathbf{A}_c(s) \in \mathbb{R}^{N_\mathrm{y} \times N_\mathrm{x}}$ and $\mathbf{B}_c(s) \in \mathbb{R}^{N_\mathrm{x} \times N_\mathrm{x}}$ after $s$ iterations.
The two matrices in the next step are calculated with new data $\mathbf{\tilde{A}}_c$ and $\mathbf{\tilde{B}}_c$ as follows:
\begin{align}
	\mathbf{A}_c(s+1) & = \mathbf{A}_c(s) + \mathbf{\tilde{A}}_c, \\
	\mathbf{B}_c(s+1) & = \mathbf{B}_c(s) + \mathbf{\tilde{B}}_c.
	\label{eq:increment}
\end{align}
As seen above, this method enables incremental updates of readout weights.
An overview of how the weights are transmitted between clients and a server in IncFed ESN is shown in Fig. \subref*{fig:incfed-esn}.

If one applies IncFed ESN to anomaly detection, a standard approach uses reconstruction error as the anomaly score.
In such a reconstruction error-based method, models are trained to reconstruct the inputs by the model output using only normal data.
Then the reconstruction errors for unknown normal input data tend to be small because of successful reconstruction, whereas those for anomaly inputs tend to be large due to reconstruction failure.

However, this error-based approach has two challenges in anomaly detection\cite{MD-RS}.
First, anomaly detection heavily relies on the accuracy of reconstruction.
Models with low reconstruction accuracy lack reliability in detecting anomalies.
Second, reconstruction error is highly vulnerable to input noise.

\section{Proposed Method}
The use of reconstruction error as an anomaly score has the aforementioned issues.
Another promising approach for anomaly detection is to leverage a deviation from a learned distribution corresponding to normal data for anomaly scoring\cite{schmidl2022anomaly,choi2021deep}.
A method following this latent distribution-based approach in reservoir computing is Mahalanobis Distance of Reservoir States (MD-RS) \cite{MD-RS}.
We introduce MD-RS for anomaly detection in Sec. \ref{sec:MD-RS} and propose a novel federated learning scheme named Incremental Federated Learning with MD-RS (IncFed MD-RS) in Sec. \ref{sec:IncFed_MD-RS}.

\subsection{Malahanobis Distance of Reservoir States (MD-RS)}
\label{sec:MD-RS}
\begin{figure}
	\centering
	\includegraphics[width=0.48\textwidth]{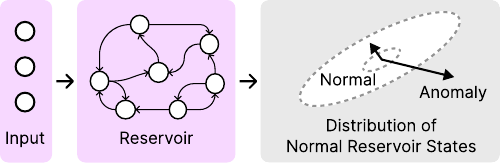}
	\caption{Overview of Mahalanobis Distance of Reservoir States.}
	\label{fig:md-rs}
\end{figure}

In MD-RS, the reservoir layer behaves as a feature converter that nonlinearly maps input temporal features into a high-dimensional feature space.
First, MD-RS approximates the reservoir state vectors for normal input data by a multivariate Gaussian distribution, estimating its mean vector and covariance matrix in the training phase.
Then, after training, MD-RS uses Mahalanobis distance between a reservoir state vector for new input data and the learned distribution as an anomaly score.
The distance denotes how far the current reservoir state vector deviates from the normal subspace.
As seen above, MD-RS is a semi-supervised learning method that requires only normal data for training, which makes this method suitable for practical scenarios with rare anomalous data.
We emphasize that this reservoir state analysis does not require a readout layer.
An overview of MD-RS is shown in Fig. \ref{fig:md-rs}.

In MD-RS, the covariance matrix of reservoir states during training is defined as follows:
\begin{equation}
	\mathbf{\Phi}_0 := \sum_{t=1}^{T_0} \mathbf{x}(t)\mathbf{x}^\top(t),
	\label{eq:covariance-matrix}
\end{equation}
where $T_0$ denotes the length of the training data.

Then the precision matrix $\mathbf{P}_\mathrm{0} \in \mathbb{R}^{N_\mathrm{x} \times N_\mathrm{x}}$ of reservoir states in the training phase is defined as follows:
\begin{equation}
	\mathbf{P}_0 := (\mathbf{\Phi}_0 + \delta \mathbf{I})^{-1},
	\label{eq:precision-matrix}
\end{equation}
where $\delta$ is a regularization parameter larger than 0, $\mathbf{I} \in \mathbb{R}^{N_\mathrm{x} \times N_\mathrm{x}}$ is an identity matrix and the $\delta \mathbf{I}$ term contributes to stabilization of inverse calculation.

In the test phase, the anomaly score for a reservoir state vector $\mathbf{x}(t)$ generated by a new input through \eqref{eq:internal-state} is computed as follows:
\begin{equation}
	D_{\rm{M}}^2(\mathbf{x}(t), \mathbf{P}_\mathrm{0}) := \mathbf{x}^\top(t) \mathbf{P}_0 \mathbf{x}(t).
	\label{eq:md-rs}
\end{equation}
Note that the above definition assumes the mean vector of the reservoir state as zero vector.
A large distance means that the input deviates much from the group of normal inputs, and indicates it is likely to be anomalous.

Notably, the precision matrix $\mathbf{P}(t)$ can be updated online with the following equation:
\begin{equation}
	\mathbf{P}(t+1) =  \mathbf{P}(t) - \frac{\mathbf{P}(t)\mathbf{x}(t)\mathbf{x}^\top(t)\mathbf{P}(t)}{1 +\mathbf{x}^\top(t)\mathbf{P}(t)\mathbf{x}(t)},
	\label{eq:online-precision}
\end{equation}
which is derived from \eqref{eq:covariance-matrix} and \eqref{eq:precision-matrix} using Woodbury formula.
The initial condition of the precision matrix is provided by
\begin{equation}
	\mathbf{P}(0) := \frac{\mathbf{I}}{\delta}.
	\label{eq:initial-precision}
\end{equation}
Due to the size of $\mathbf{x}(t)$, the time complexity in \eqref{eq:online-precision} is $O(N_\mathrm{x}^2)$, while the time complexity in \eqref{eq:precision-matrix} is $O(N_\mathrm{x}^3)$.
Therefore, the iterative updating in \eqref{eq:online-precision} is more computationally efficient than \eqref{eq:precision-matrix}, given that the precision matrix is updated every time new data is available.
If it is enough for the precision matrix to be calculated only once for a certain data sequence of length $T_0 \gg N_\mathrm{x}$, \eqref{eq:precision-matrix} is more efficient than \eqref{eq:online-precision} because the total time complexity is $O(N_\mathrm{x}^3)$ for \eqref{eq:precision-matrix} and $O(N_\mathrm{x}^2 T_0)$ for \eqref{eq:online-precision}.

\subsection{Incremental Federated Learning with MD-RS}
\label{sec:IncFed_MD-RS}
We propose Incremental Federated Learning with Mahalanobis Distance of Reservoir States (IncFed MD-RS).

First, the client covariance matrix $\mathbf{\Phi}_c$ at the client $c \in C$, interpreted as a local model in the federated learning context, is defined as follows:
\begin{equation}
	\mathbf{\Phi}_c := \sum_{t=1}^{T_0} \mathbf{x}_c(t)\mathbf{x}_c^\top(t),
	\label{eq:covariance-matrix-federated}
\end{equation}
where $\mathbf{x}_c(t)$ denotes the reservoir state vector at time step $t$ on client $c$ and $T_0$ is the length of client $c$'s training data.
During training, clients update their own precision matrix incrementally following \eqref{eq:online-precision}, while computing the covariance matrix of training data as in \eqref{eq:covariance-matrix} in order to construct the global model.

After all clients send their covariance matrices to the server, the server computes a global covariance matrix by summing those matrices as follows:
\begin{equation}
	\mathbf{\Phi}_g := \sum_{c \in C} \mathbf{\Phi}_c.
	\label{eq:global-covariance-matrix}
\end{equation}

In this way, we can obtain global precision matrix $\mathbf{P}_g$ by computing the inverse matrix of the covariance matrix as follows:
\begin{equation}
	\mathbf{P}_g := (\mathbf{\Phi}_g + \delta\mathbf{I})^{-1},
	\label{eq:global-precision-matrix}
\end{equation}
where the term $\delta\mathbf{I}$ works as a regularization to stabilize inverse calculation in the same way as \eqref{eq:precision-matrix}.
Note that the global precision matrix in \eqref{eq:global-precision-matrix} is mathematically equivalent to those in \eqref{eq:precision-matrix} and \eqref{eq:online-precision} given that all data are available to the server.

Based on the global precision matrix $\mathbf{P}_g$, squared Mahalanobis distance $D_\mathrm{M}^2(\mathbf{x}(t), \mathbf{P}_g)$ is calculated during testing.

Similarly to IncFed ESN, the client covariance matrix $\mathbf{\Phi}_c(s)$ after $s$ iterations can be updated incrementally.
After new data recognized as normal are added, a new covariance matrix is calculated with the existing covariance matrix.
Given that the covariance matrix of new data in client $c$ be $\mathbf{\tilde{\Phi}}_c$, the client $c$ sums the matrices as follows:
\begin{equation}
	\mathbf{\Phi}_c(s+1) = \mathbf{\Phi}_c(s) + \mathbf{\tilde{\Phi}}_c
	\label{eq:covariance-matrix-incremental}
\end{equation}
as in \eqref{eq:covariance-matrix}.
An overview of how the covariance and precision matrices are transmitted between clients and the server in IncFed MD-RS is shown in Fig. \subref*{fig:incfed-mdrs}.

As seen above, our method enables an optimal aggregation of the local models of all clients and incremental updates of global models.

\subsection{Subsampling Reservoir Nodes}

In IncFed ESN, the matrices transferred from each client to the server are $\mathbf{A}_c$ and $\mathbf{B}_c$ in \eqref{eq:Ac} and \eqref{eq:Bc}, which are the product of the target output matrix with the reservoir state matrix, and the product of reservoir state matrix with its transpose, respectively.
The number of values from each client to the server is $N_\mathrm{x}^2 + N_\mathrm{x}N_\mathrm{y}$.
Unlike IncFed ESN, IncFed MD-RS necessitates just the covariance matrix of reservoir states $\mathbf{\Phi}_c$.
Therefore, the number of values transmitted from each client to the server is $N_\mathrm{x}^2$.
To reduce communication costs further and improve computational efficiency, we adopt a new approach, subsampling.

The authors in \cite{MD-RS} introduce subsampling of reservoir nodes into MD-RS, which uses a subset of reservoir nodes to calculate Mahalanobis distance instead of whole reservoir nodes.
With subsampling size $\tilde{N}_\mathrm{x} < N_\mathrm{x}$, $\tilde{N}_\mathrm{x}$ nodes are randomly selected from the $N_\mathrm{x}$ reservoir nodes.
Then a subsampled reservoir state vector, denoted by $\tilde{\mathbf{x}}(t) \in \mathbb{R}^{\tilde{N}_\mathrm{x}}$, is used in \eqref{eq:covariance-matrix}, \eqref{eq:md-rs}, \eqref{eq:online-precision}, and \eqref{eq:covariance-matrix-federated} instead of the whole reservoir state vector $\mathbf{x}(t)$.
The time complexity is $O(\tilde{N}_\mathrm{x}^3)$ in \eqref{eq:precision-matrix}, and $O(\tilde{N}_\mathrm{x}^2)$ in \eqref{eq:online-precision} for each time step, which leads to further computational efficiency than that without subsampling.

IncFed MD-RS can adopt this approach likewise, and notably, it contributes to communication cost reduction as well as computational efficiency.
As a result, the number of values transmitted from a client to the server is $\tilde{N}_\mathrm{x}^2$ because the matrix sent from a client to a server is $\mathbf{\tilde{\Phi}}_c \in \mathbb{R}^{\tilde{N}_\mathrm{x} \times \tilde{N}_\mathrm{x}}$.

\section{Experiments}
\subsection{Datasets}
We conduct experiments using the following three real-world time series datasets for anomaly detection, contained in the federated time series anomaly detection benchmark, named FedTADBench\cite{FedTADBench}.

\begin{itemize}
	\item \textbf{SMD}. Server Machine Dataset (SMD)\cite{SMD} is a 5-week-long dataset collected from a large internet company and made up of data from 28 different machines.
	\item \textbf{SMAP}. Soil Moisture Active Passive satellite (SMAP)\cite{SMAP} is an expert-labeled telemetry anomaly data from NASA.
	\item \textbf{PSM}. Pooled Server Metrics (PSM)\cite{PSM} is a dataset collected internally from multiple application server nodes at eBay.
\end{itemize}
The details of the datasets we introduced above are shown in Table \ref{tab:dataset-details}.

Unlike the other datasets, each time series in SMAP consists of only one sensor channel, and the other channels are one-hot-encoded commands given to each entity.
For SMAP, we use only its telemetry sensor channel as input to models.

The time series contained in PSM is just one sequence, while SMD and SMAP consist of multiple time series.
In order to conduct federated learning experiments in which training data are distributed among clients, we divide the training time series data by the number of clients.
Following \cite{FedTADBench}, the number of clients is set to 24.

\begin{table}[!t]
	\renewcommand{\arraystretch}{1.3}
	\caption{
		Dataset Characteristics.
		TS means the number of time series in the datasets.
		Dims means the dimension of data.
		Avg. Tr. Len. represents the average training length.
		Avg. Te. Len. means average test length.
		Anomalies mean the percentage of anomalous data.
		The unit for Anomalies is \%.
	}
	\label{tab:dataset-details}
	\centering
	\begin{tabular}{lccccc}
		\toprule
		Dataset & TS & Dims & Avg. Tr. Len. & Avg. Te. Len. & Anomalies \\
		\midrule
		SMD     & 28 & 38   & 25300         & 25300         & 4.16      \\
		SMAP    & 55 & 1    & 2560          & 8073          & 13.1      \\
		PSM     & 1  & 25   & 132481        & 87841         & 27.8      \\
		\bottomrule
	\end{tabular}
\end{table}

\subsection{Evaluation Metrics}

We choose the following four threshold-independent evaluation metrics for time series anomaly detection:
\begin{itemize}
	\item \textbf{AUC-ROC}. The Area Under Receiver Operating Characteristic Curve (AUC-ROC) measures the trade-off between False Positive Rate and True Positive Rate across different thresholds.
	\item \textbf{AUC-PR}. The Area Under Precision-Recall Curve (AUC-PR) measures the trade-off between Precision and Recall across different thresholds.
	\item \textbf{VUS-PR}. The Volume Under the Surface of Precision-Recall Curve (VUS-PR)\cite{VUS-PR} aims to overcome issues such as labeling consistency and the impact of time lags on anomaly scores.
	      Incorporating a tolerance buffer around outlier boundaries and adopting continuous values over binary labels resolve the issues.
	\item \textbf{PATE}. Proximity-Aware Time series anomaly Evaluation (PATE)\cite{PATE} computes a weighted version of AUC-PR with a proximity-based weighting approach.
	      The proximity-based weighting approach enables a detailed assessment of both early and delayed detections and even onset response time by integrating buffer zones around anomaly events.
\end{itemize}

AUC-ROC and AUC-PR are widely used evaluation metrics.
However, it is argued that both of the evaluation metrics fall short in time series contexts due to not considering the order of the data points and the temporal correlation between them.
The two other metrics, VUS-PR and PATE, have been proposed to resolve the issue.
In short, both metrics introduce a kind of buffer around anomaly events, and thus they integrate the nature of the time series into metric scores.
We set the buffer size at the default value in PATE while VUS-PR is parameter-free.

\begin{table}[!t]
	\renewcommand{\arraystretch}{1.3}
	\caption{
		Average scores of the evaluation metrics, AUC-ROC, AUC-PR, VUS-PR, and PATE, over time series in test data in Server Machine Dataset (SMD) for each method.
		We highlight the best value in bold and underline the second-best value for centralized and federated methods, respectively. "Central." represents centralized.
	}
	\label{tab:SMD}
	\centering
	\begin{tabular}{llcccc}
		\toprule
		                                                      &                & AUC-ROC           & AUC-PR            & VUS-PR            & PATE              \\
		\midrule
		                                                      & TranAD         & \underline{0.700} & \underline{0.282} & \underline{0.283} & \underline{0.313} \\
		                                                      & LSTM-AE        & 0.610             & 0.209             & 0.220             & 0.231             \\
		\multirow{-3}{*}{\rotatebox[origin=c]{90}{Central.}}  & MD-RS          & \textbf{0.852}    & \textbf{0.442}    & \textbf{0.488}    & \textbf{0.496}    \\
		\midrule
		                                                      & FedAvg TranAD  & 0.665             & 0.247             & \underline{0.261} & 0.276             \\
		                                                      & FedAvg LSTM-AE & 0.541             & 0.189             & 0.184             & 0.208             \\
		                                                      & IncFed ESN-SRE & 0.619             & 0.197             & 0.168             & 0.222             \\
		                                                      & FedAvg MD-RS   & \underline{0.675} & \underline{0.271} & 0.237             & \underline{0.300} \\
		\multirow{-5}{*}{\rotatebox[origin=c]{90}{Federated}} & IncFed MD-RS   & \textbf{0.852}    & \textbf{0.442}    & \textbf{0.488}    & \textbf{0.496}    \\
		\bottomrule
	\end{tabular}
\end{table}

\begin{figure}[!t]
	\centering
	\subfloat[AUC-ROC Distribution for SMD]{\includegraphics[width=0.24\textwidth]{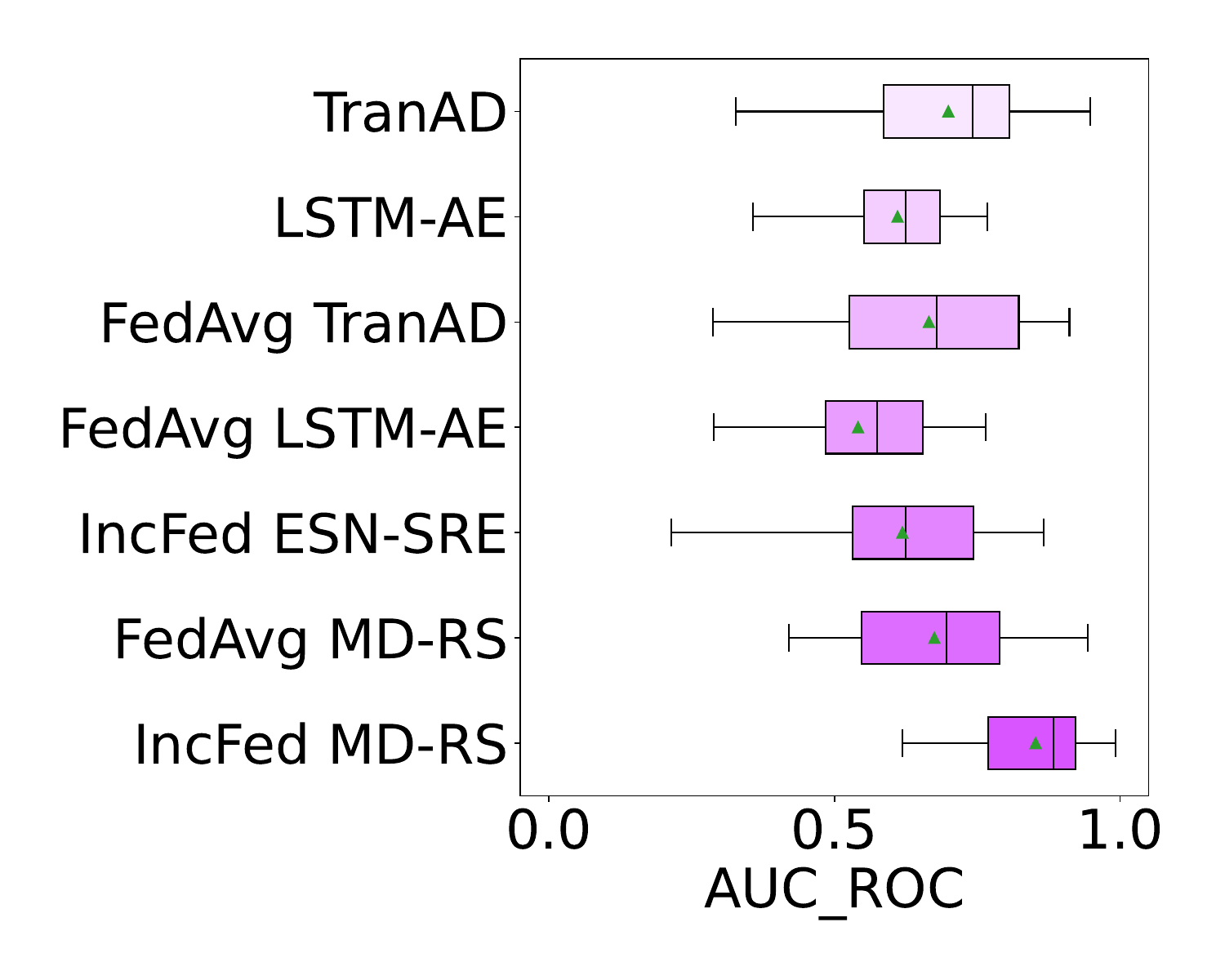}\label{SMD-AUC-ROC}}
	\hfil
	\subfloat[AUC-PR Distribution for SMD]{\includegraphics[width=0.24\textwidth]{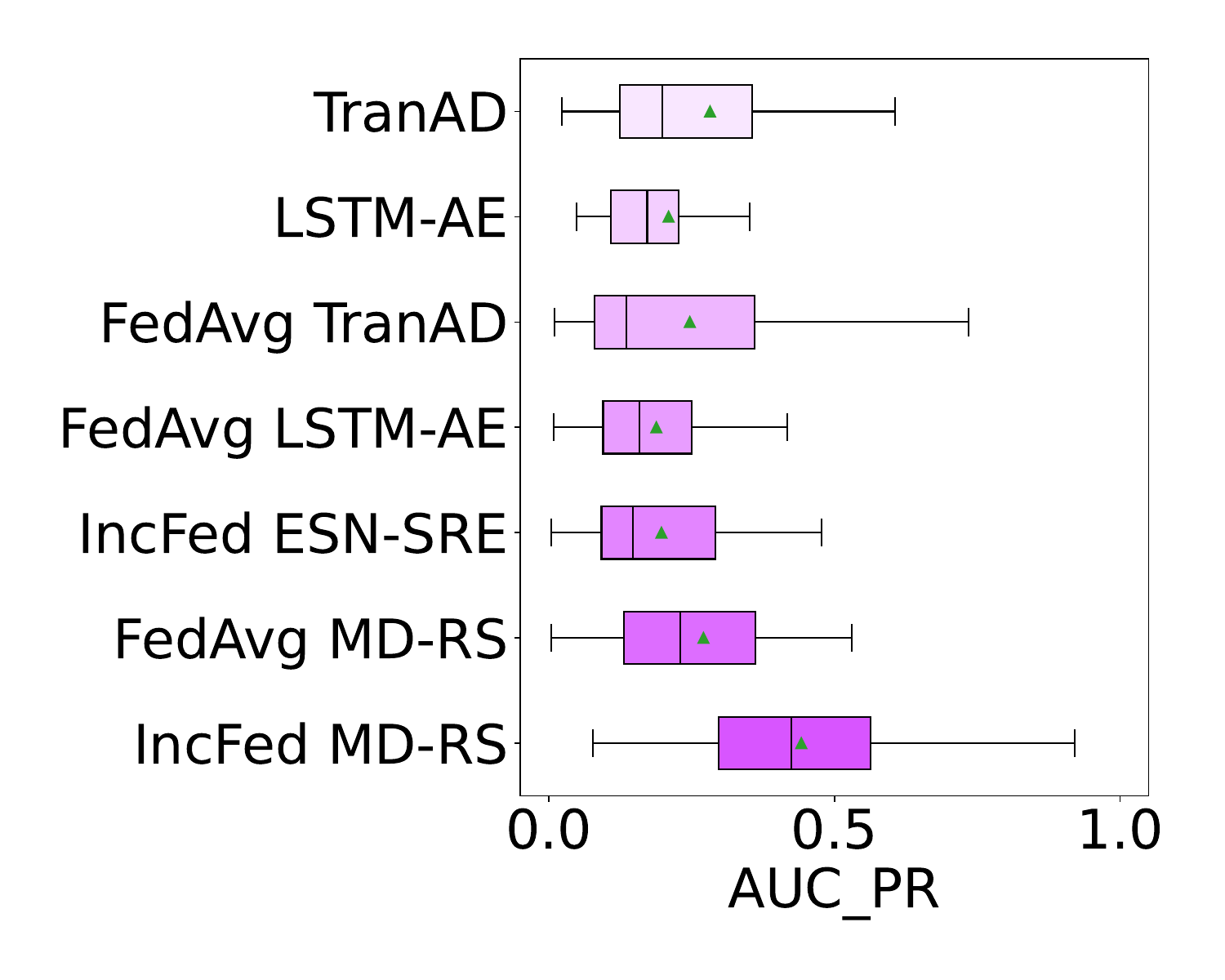}\label{SMD-AUC-PR}}
	\hfil
	\subfloat[VUS-PR Distribution for SMD]{\includegraphics[width=0.24\textwidth]{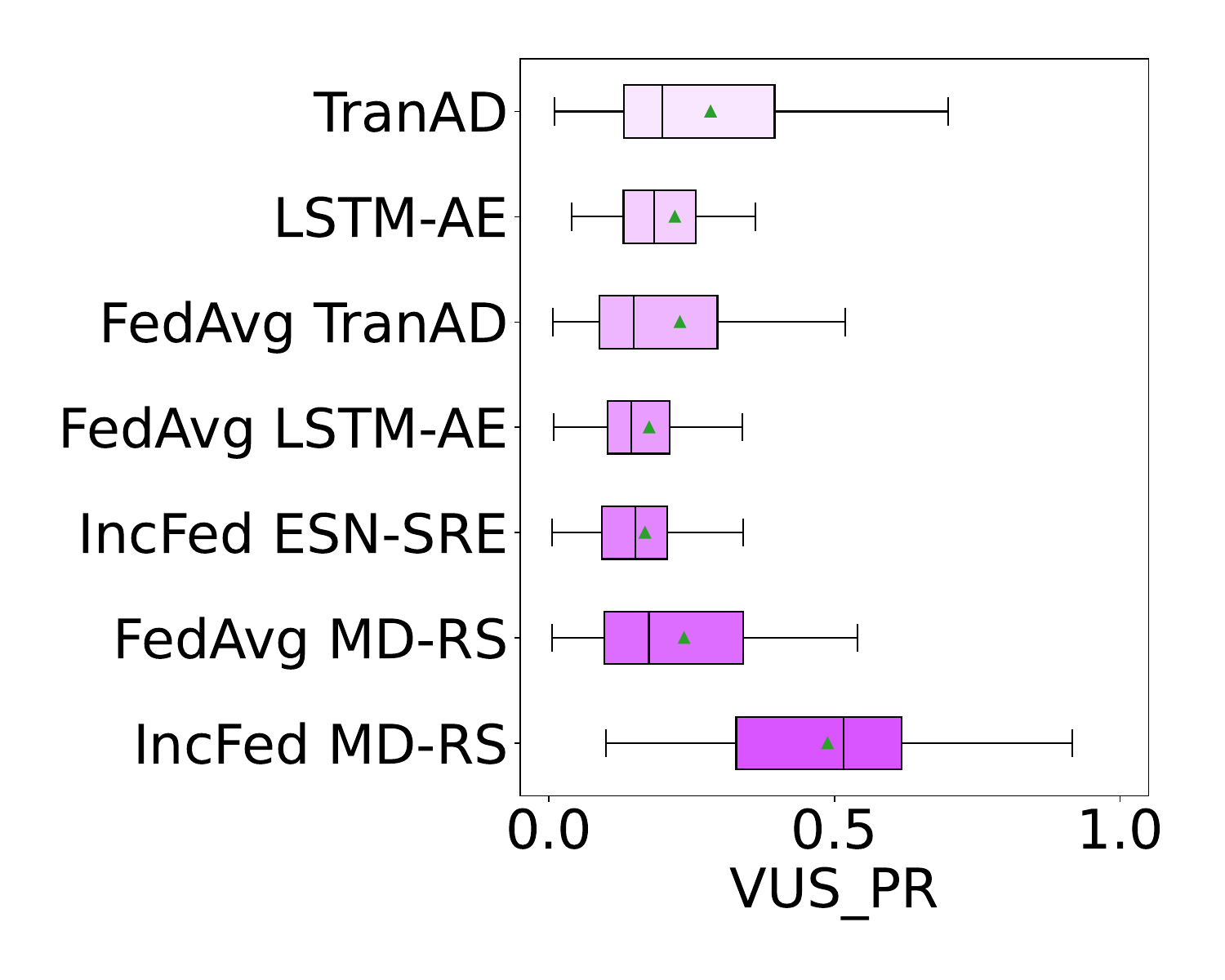}\label{SMD-vus_pr}}
	\hfil
	\subfloat[PATE Distribution for SMD]{\includegraphics[width=0.24\textwidth]{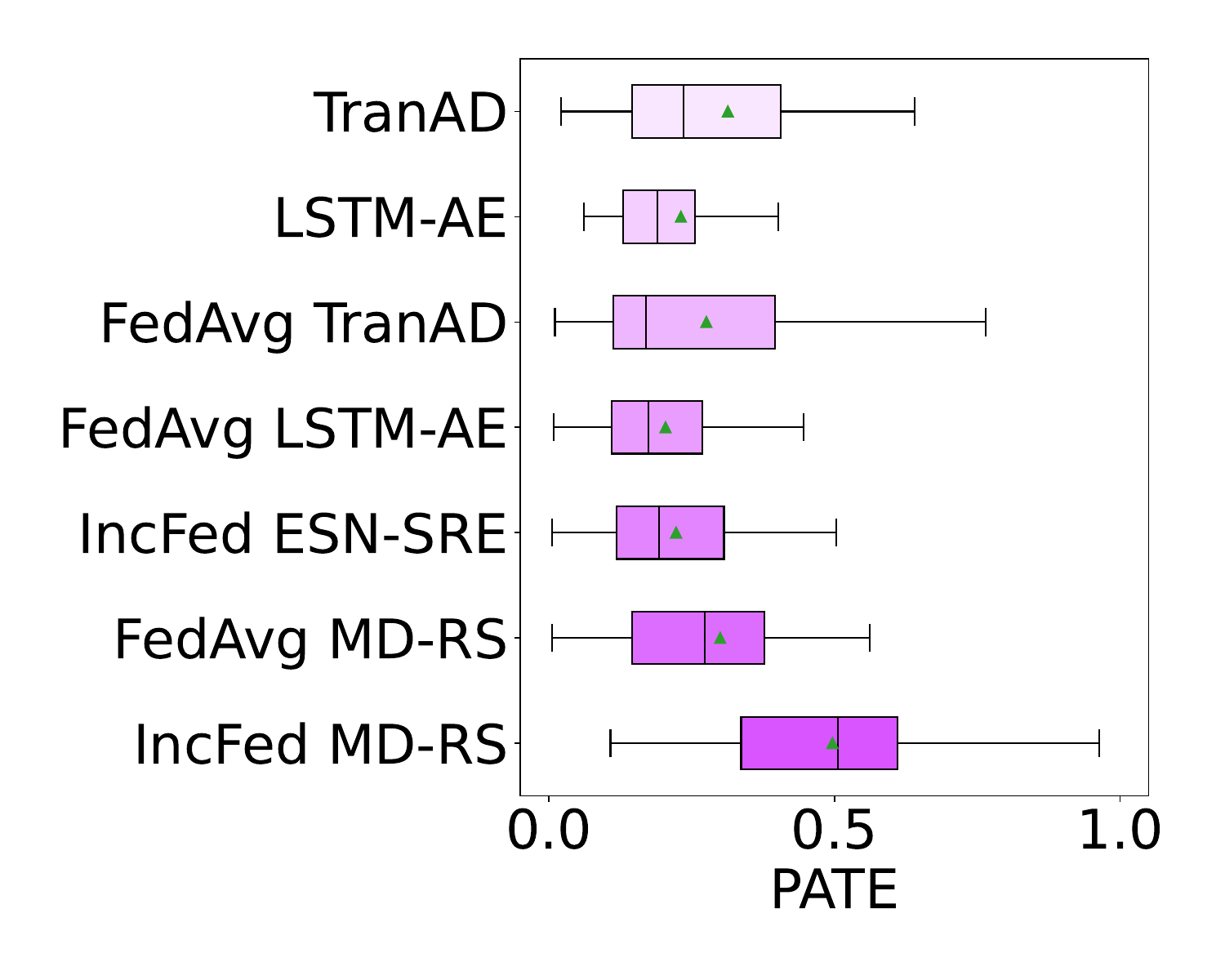}\label{SMD-PATE}}
	\hfil
	\caption{
		Distributions of the evaluation metric scores over time series in the test dataset in Server Machine Dataset (SMD).
		The triangle marks present the average evaluation metric scores.
	}
	\label{fig:evaluation-metric-distribution-for-SMD}
\end{figure}

\subsection{Experimental Settings}

For the sake of comparison, we apply some federation methods to the following machine learning methods:

\begin{itemize}
	\item \textbf{TranAD}. Transformer network for anomaly detection (TranAD)\cite{tuli2022tranad} is a reconstruction error-based method and achieves state-of-the-art performance in anomaly detection.
	\item \textbf{LSTM-AE}. Long Short-Term Memory Auto-Encoder (LSTM-AE)\cite{LSTM-ED} is a combination method of LSTM and Auto Encoders to reconstruct input data.
	\item \textbf{ESN-SRE}. This method reconstructs input data with ESN and uses reconstruction errors, such as Squared Reconstruction Error (SRE) as anomaly scores\cite{JunyaKato2024}.
\end{itemize}
For the two deep learning-based methods: TranAD and LSTM-AE, we apply FedAvg to those, denoted by FedAvg TranAD and FedAvg LSTM-AE.
For ESN-SRE, we adopt squared reconstruction errors as anomaly scores in IncFed ESN, denoted by IncFed ESN-SRE.
Moreover, we compare IncFed MD-RS with FedAvg-applied MD-RS, denoted by FedAvg MD-RS, centralized TranAD, LSTM-AE, and MD-RS for reference.

We evaluate the performance of the models for each time series in the test dataset and calculate the mean value of those evaluation metric scores.
There are 28, 55, and 1 scores over evaluation metrics for SMD, SMAP, and PSM, respectively.

The hyperparameters for IncFed MD-RS and FedAvg MD-RS are the same across all datasets with the number of reservoir nodes $N_\mathrm{x} = 500$, the subsampling size $\tilde{N}_\mathrm{x} = 200$, leaking rate $\alpha = 1.0$, spectral radius $\rho = 0.95$, input scaling 0.001, and regularization parameter $\delta = 0.0001$.

\begin{table}[!t]
	\renewcommand{\arraystretch}{1.3}
	\caption{
		Average scores of the evaluation metrics, AUC-ROC, AUC-PR, VUS-PR, and PATE, over test time series in Soil Moisture Active Passive satellite (SMAP) for each method.
		We highlight the best value in bold and underline the second-best value for centralized and federated methods, respectively. "Central." represents centralized.
	}
	\label{tab:SMAP}
	\centering
	\begin{tabular}{llcccc}
		\toprule
		                                                      &                & AUC-ROC           & AUC-PR            & VUS-PR            & PATE              \\
		\midrule
		                                                      & TranAD         & \underline{0.567} & 0.280             & 0.262             & 0.335             \\
		                                                      & LSTM-AE        & 0.546             & \underline{0.304} & \underline{0.296} & \underline{0.347} \\
		\multirow{-3}{*}{\rotatebox[origin=c]{90}{Central.}}  & MD-RS          & \textbf{0.659}    & \textbf{0.475}    & \textbf{0.500}    & \textbf{0.500}    \\
		\midrule
		                                                      & FedAvg TranAD  & 0.569             & 0.307             & 0.287             & 0.370             \\
		                                                      & FedAvg LSTM-AE & 0.574             & 0.324             & 0.319             & 0.368             \\
		                                                      & IncFed ESN-SRE & 0.635             & 0.374             & 0.404             & 0.395             \\
		                                                      & FedAvg MD-RS   & \underline{0.641} & \underline{0.408} & \underline{0.436} & \underline{0.430} \\
		\multirow{-5}{*}{\rotatebox[origin=c]{90}{Federated}} & IncFed MD-RS   & \textbf{0.659}    & \textbf{0.475}    & \textbf{0.500}    & \textbf{0.500}    \\
		\bottomrule
	\end{tabular}
\end{table}

\begin{figure}[!t]
	\centering
	\subfloat[AUC-ROC Distribution for SMAP]{\includegraphics[width=0.24\textwidth]{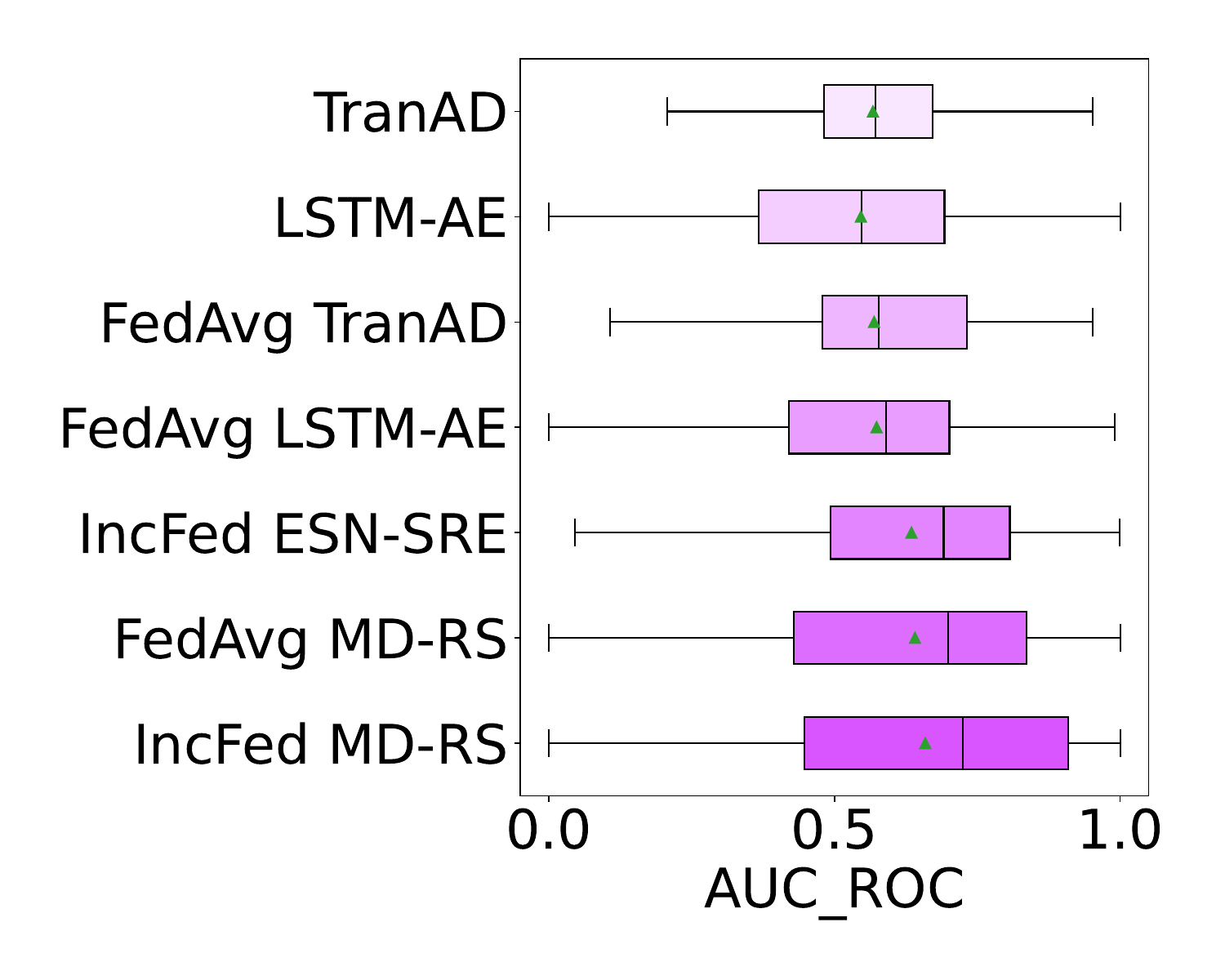}\label{SMAP-AUC-ROC}}
	\hfil
	\subfloat[AUC-PR Distribution for SMAP]{\includegraphics[width=0.24\textwidth]{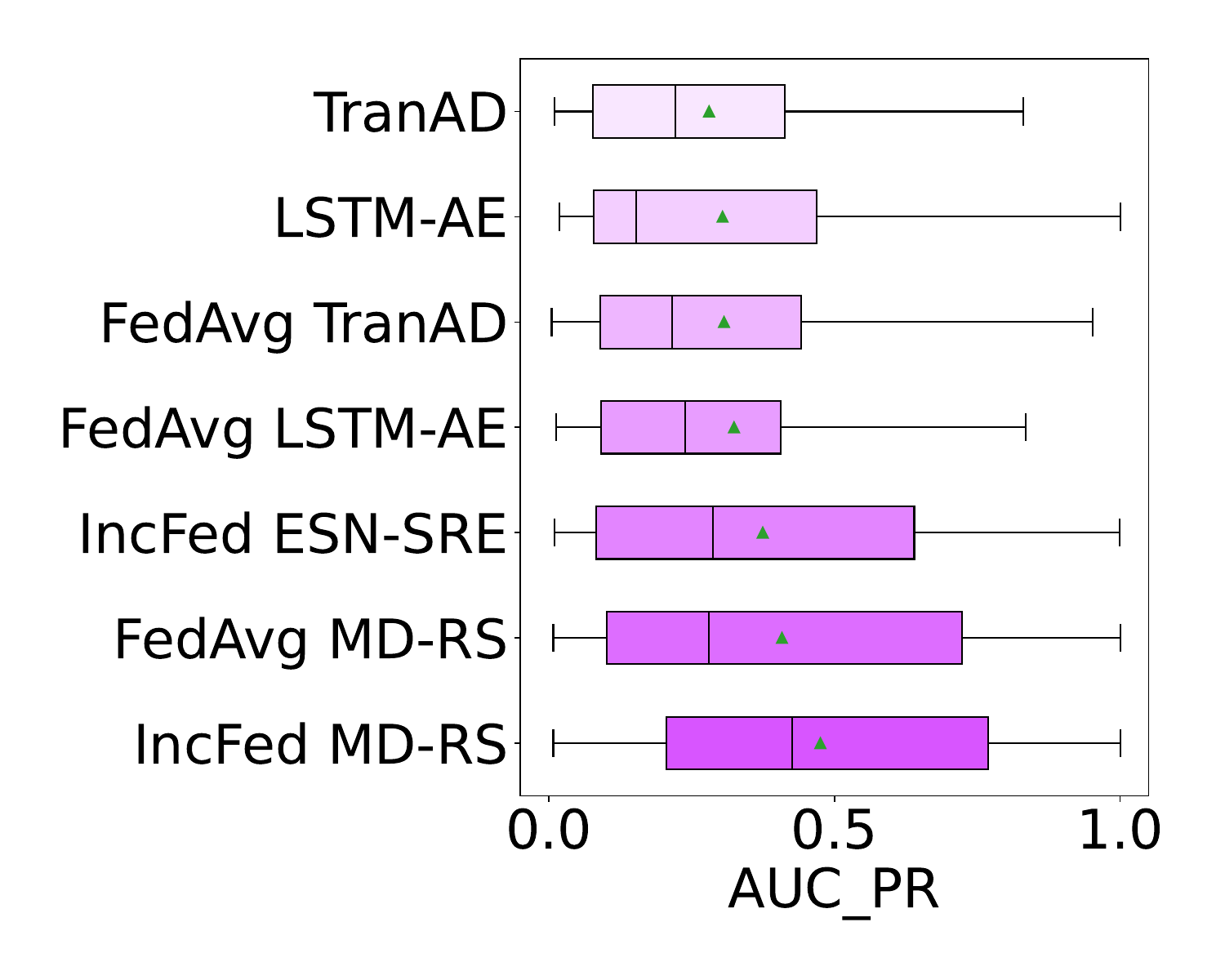}\label{SMAP-AUC-PR}}
	\hfil
	\subfloat[VUS-PR Distribution for SMAP]{\includegraphics[width=0.24\textwidth]{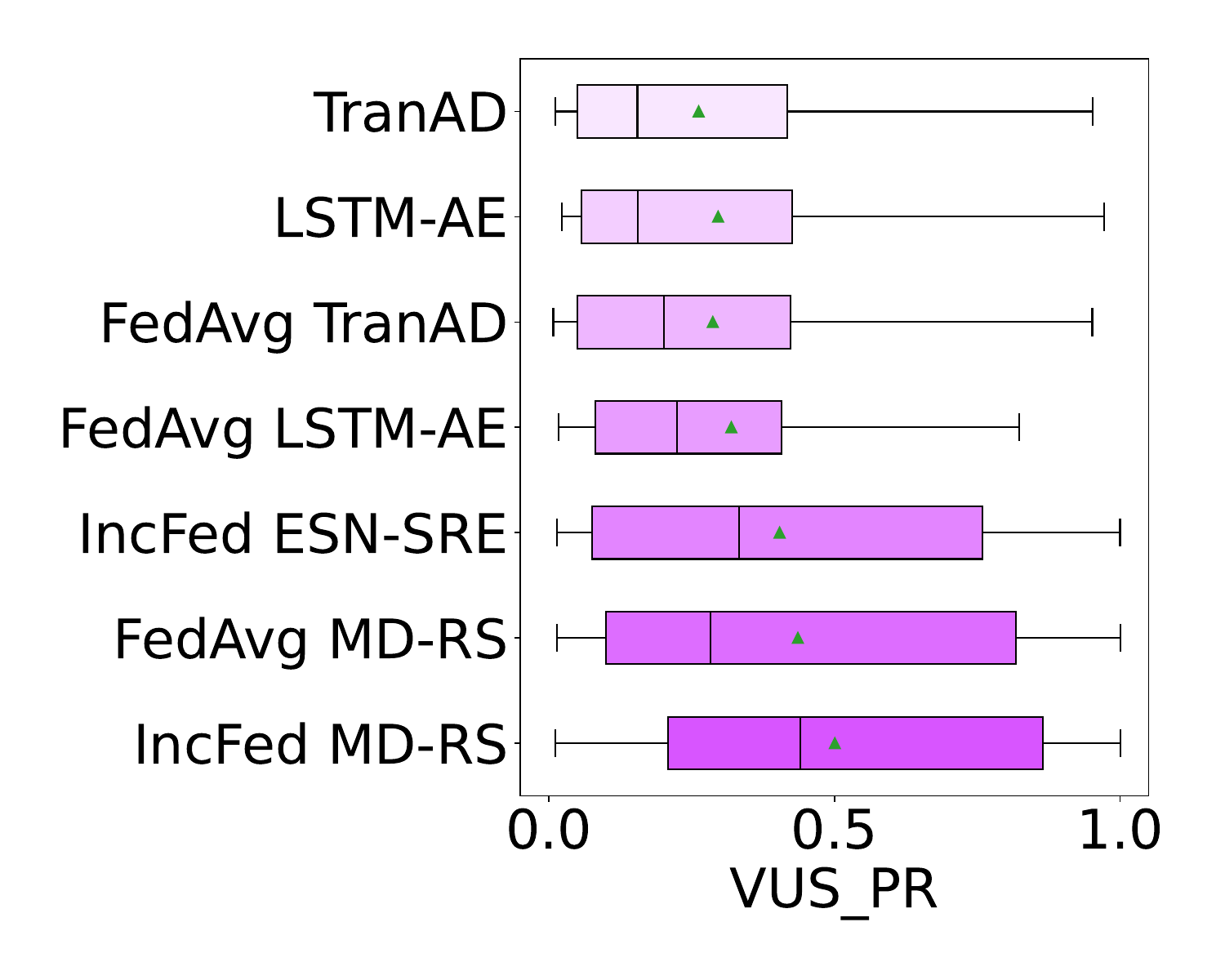}\label{SMAP-vus-pr}}
	\hfil
	\subfloat[PATE Distribution for SMAP]{\includegraphics[width=0.24\textwidth]{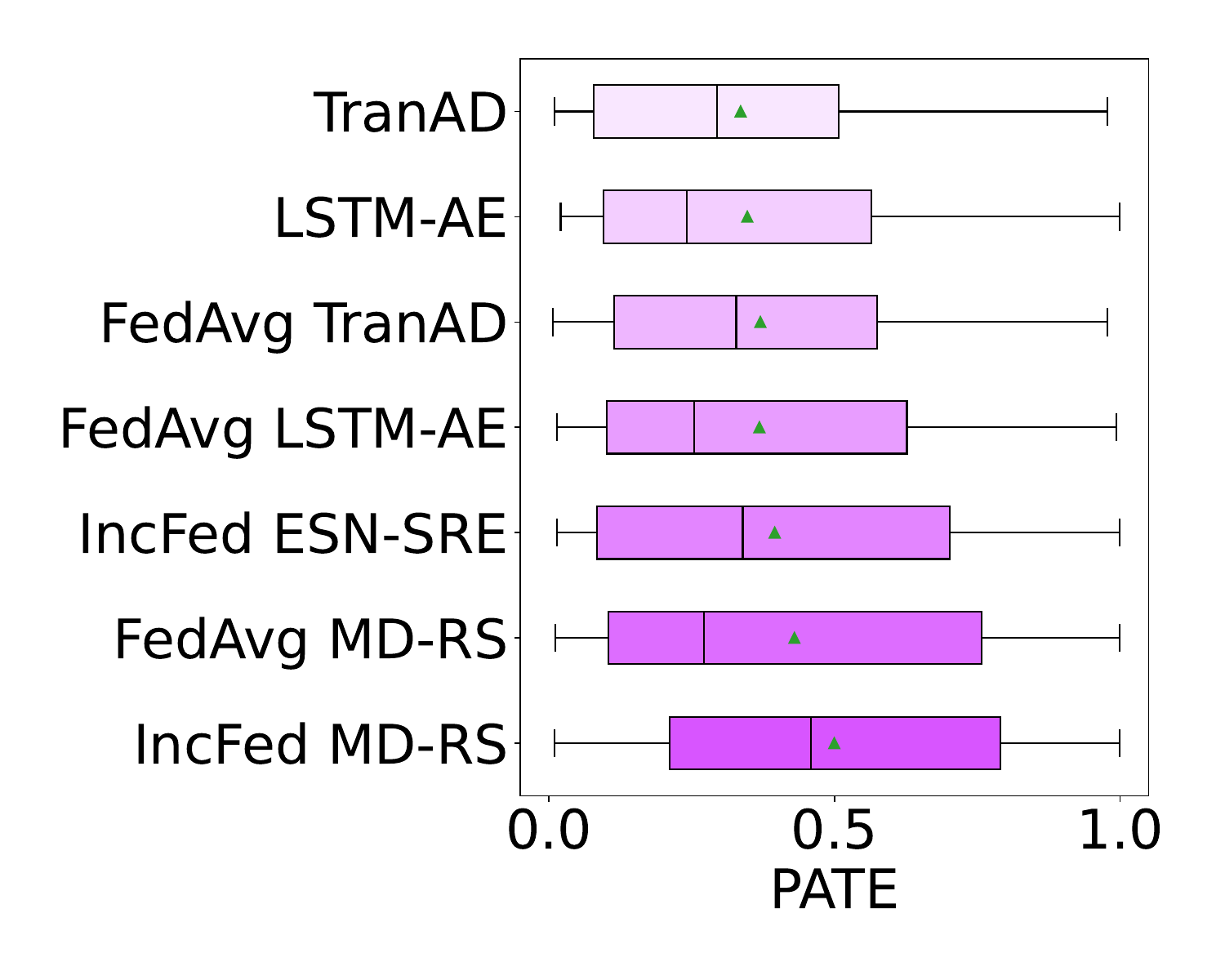}\label{SMAP-pate}}
	\caption{
		Distributions of the evaluation metric scores over time series in the test dataset in Soil Moisture Active Passive satellite (SMAP).
		The triangle marks present the average evaluation metric scores.
	}
	\label{fig:evaluation-metric-distribution-for-SMAP}
\end{figure}

\subsection{Results}

Table \ref{tab:SMD} shows the average values of the evaluation metrics, AUC-ROC, AUC-PR, VUS-PR, and PATE, for each test time series on SMD.
Fig. \ref{fig:evaluation-metric-distribution-for-SMD} shows the distribution of those evaluation metrics scores.
The triangle marks represent the average values for each metric, corresponding to the values in Table \ref{tab:SMD}.
The result shows that IncFed MD-RS significantly outperforms other methods across all evaluation metrics.
These results can be attributed to data heterogeneity between time series in the dataset.
Each time series in SMD has different characteristics from one another.
The result suggests that IncFed MD-RS is robust to data heterogeneity between clients, while deep learning models such as TranAD and LSTM-AE are unable to effectively train the models to reconstruct input data.

Table \ref{tab:SMAP} represents the average values of each evaluation metric on SMAP.
Fig. \ref{fig:evaluation-metric-distribution-for-SMAP} shows the distribution of performance for each time series in the test dataset.
The result shows that IncFed MD-RS outperforms other methods in terms of all evaluation metrics.
Similarly to SMD, the time series in SMAP have different characteristics, which leads to these results.

\begin{table}[!t]
	\renewcommand{\arraystretch}{1.3}
	\caption{
		Scores of the evaluation metrics, AUC-ROC, AUC-PR, VUS-PR, and PATE, for test time series in Pooled Server Metrics (PSM) for each method.
		We highlight the best value in bold and underline the second-best value for centralized and federated methods, respectively. "Central." represents centralized.
	}
	\label{tab:PSM}
	\centering
	\begin{tabular}{llcccc}
		\toprule
		                                                      &                & AUC-ROC           & AUC-PR            & VUS-PR            & PATE              \\
		\midrule
		                                                      & TranAD         & \textbf{0.852}    & \textbf{0.702}    & \underline{0.669} & \textbf{0.716}    \\
		                                                      & LSTM-AE        & 0.710             & 0.575             & 0.592             & 0.587             \\
		\multirow{-3}{*}{\rotatebox[origin=c]{90}{Central.}}  & MD-RS          & \underline{0.836} & \underline{0.648} & \textbf{0.680}    & \underline{0.671} \\
		\midrule
		                                                      & FedAvg TranAD  & \underline{0.763} & 0.594             & 0.557             & 0.612             \\
		                                                      & FedAvg LSTM-AE & 0.721             & \underline{0.622} & \underline{0.602} & \underline{0.634} \\
		                                                      & IncFed ESN-SRE & 0.614             & 0.440             & 0.382             & 0.460             \\
		                                                      & FedAvg MD-RS   & 0.675             & 0.538             & 0.498             & 0.561             \\
		\multirow{-5}{*}{\rotatebox[origin=c]{90}{Federated}} & IncFed MD-RS   & \textbf{0.836}    & \textbf{0.648}    & \textbf{0.680}    & \textbf{0.671}    \\
		\bottomrule
	\end{tabular}
\end{table}

\begin{figure}[!t]
	\centering
	\subfloat[]{\includegraphics[width=0.24\textwidth]{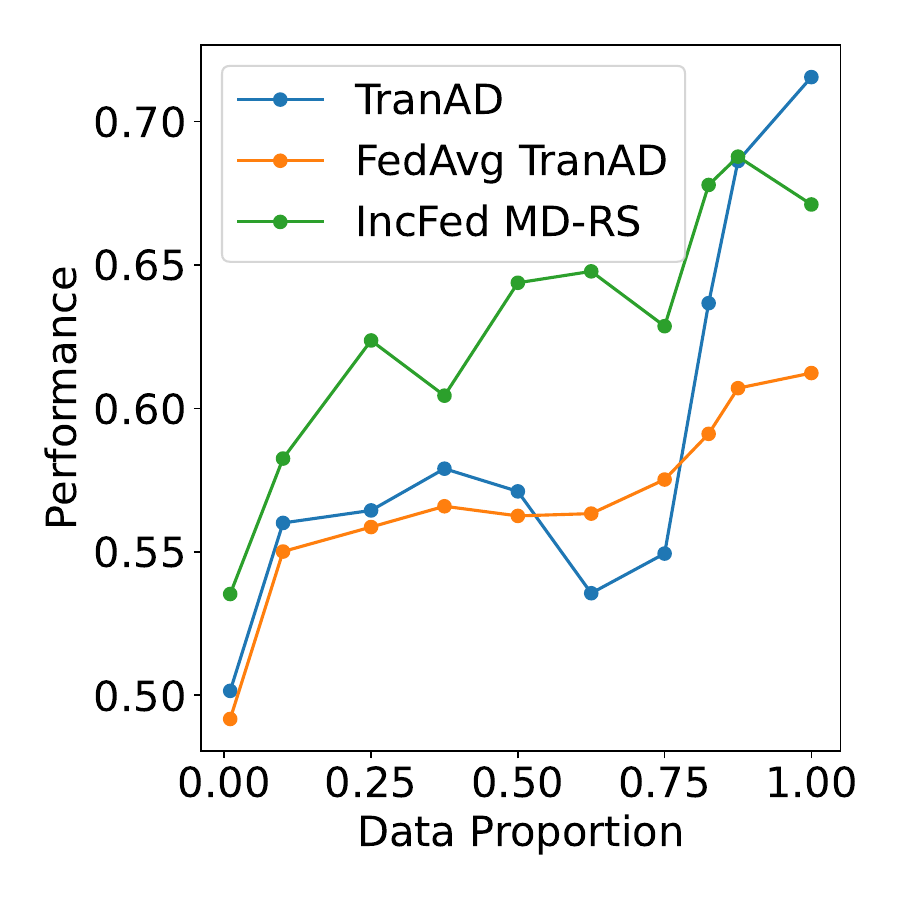}\label{fig:vary_PSM_size}}
	\hfil
	\subfloat[]{\includegraphics[width=0.24\textwidth]{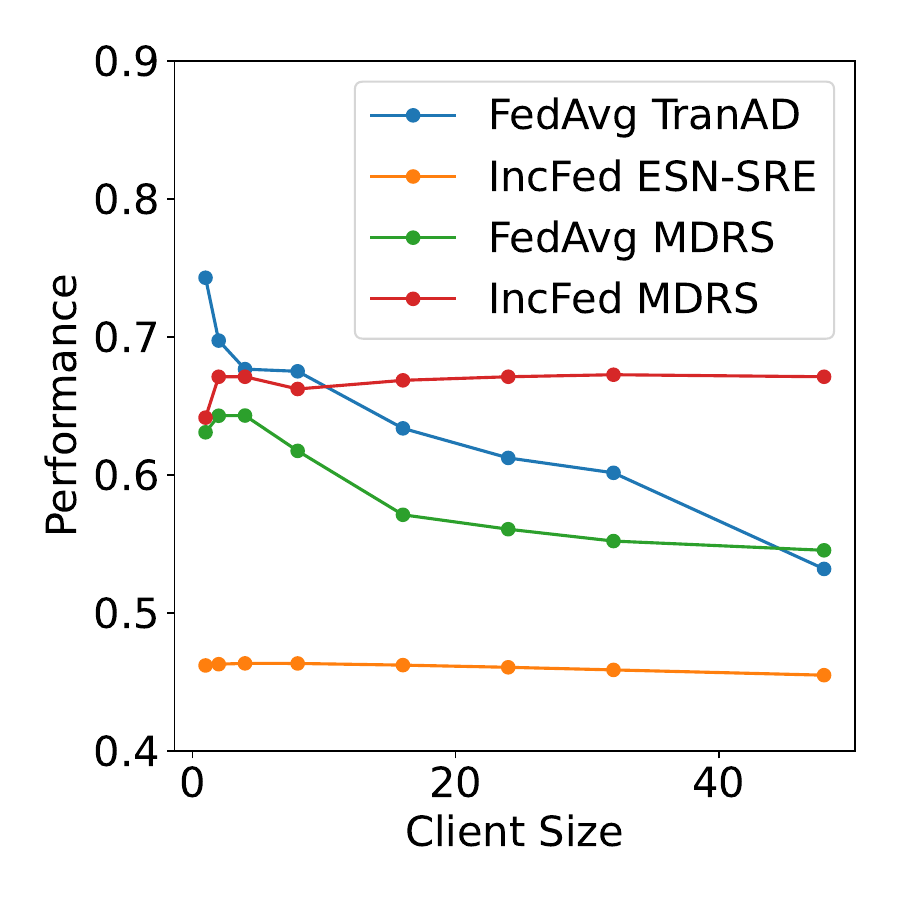}\label{fig:vary_client_size}}
	\caption{
		Various performance comparisons on Pooled Server Metrics (PSM).
		(a) Performance of each method for different training data proportions.
		(b) Performance of each method with the varying number of clients.
	}
	\label{fig:experiment-in-PSM}
\end{figure}

Table \ref{tab:PSM} shows the evaluation metric scores on PSM.
For PSM, the centralized TranAD achieves the best performance among all anomaly detection methods while its FedAvg variant shows significantly lower performance.
This result can be explained by two factors: the large number of clients and PSM's nature as a single time series with a large amount of data.

Fig. \subref*{fig:vary_PSM_size} shows how the length of training data in PSM affects the performance of two top-performing models and its variant in Table \ref{tab:PSM}: TranAD, IncFed MD-RS, and FedAvg TranAD.
In the experiment, we vary the length of training data by using a proportion of the whole data and calculate PATE scores.
With varying data proportions, we ensure that the training data used in the centralized and decentralized methods are the same.
Specifically, centralized TranAD uses a specified proportion of data taken from the beginning, while the two decentralized methods, FedAvg TranAD and IncFed MD-RS, first clip data from the beginning and then divide them so that all clients have an equal amount of data.
As seen in Fig. \subref*{fig:vary_PSM_size}, IncFed MD-RS outperforms TranAD and FedAvg TranAD at all points except for the data proportion 1.0.
This result shows that IncFed MD-RS performs well with relatively short training data compared to the two deep learning-based methods.

\begin{figure}[!t]
	\centering
	\subfloat[]{\includegraphics[width=0.24\textwidth]{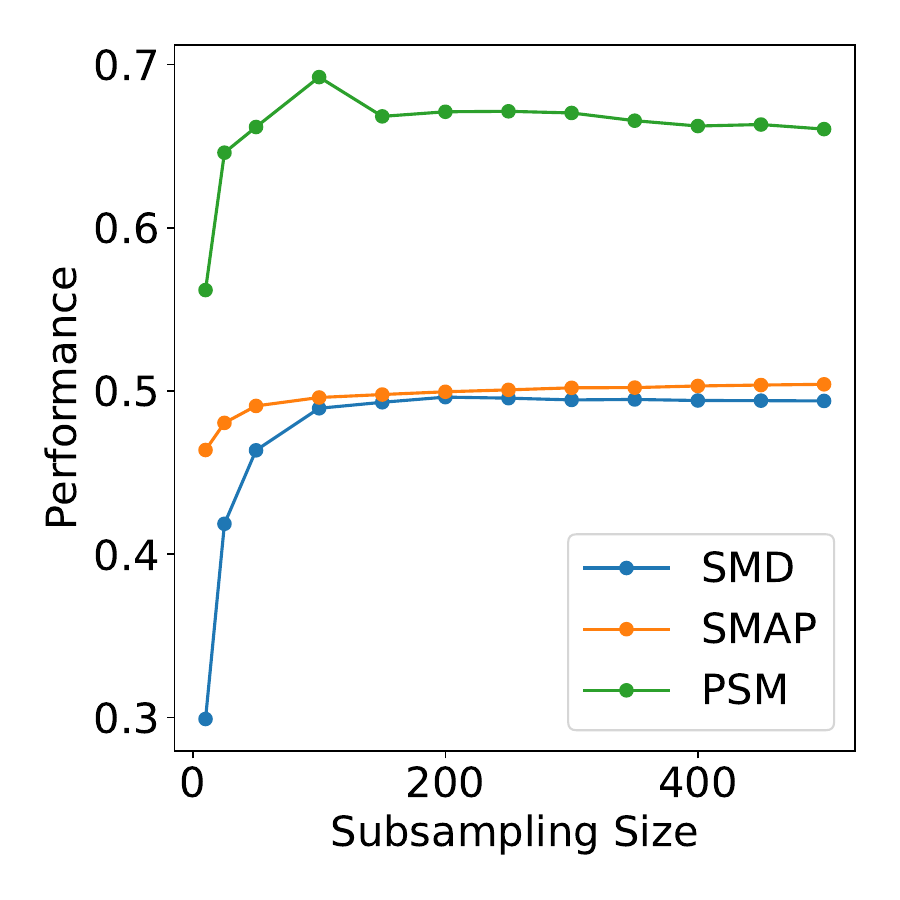}\label{fig:subsampling-performance}}
	\hfil
	\subfloat[]{\includegraphics[width=0.24\textwidth]{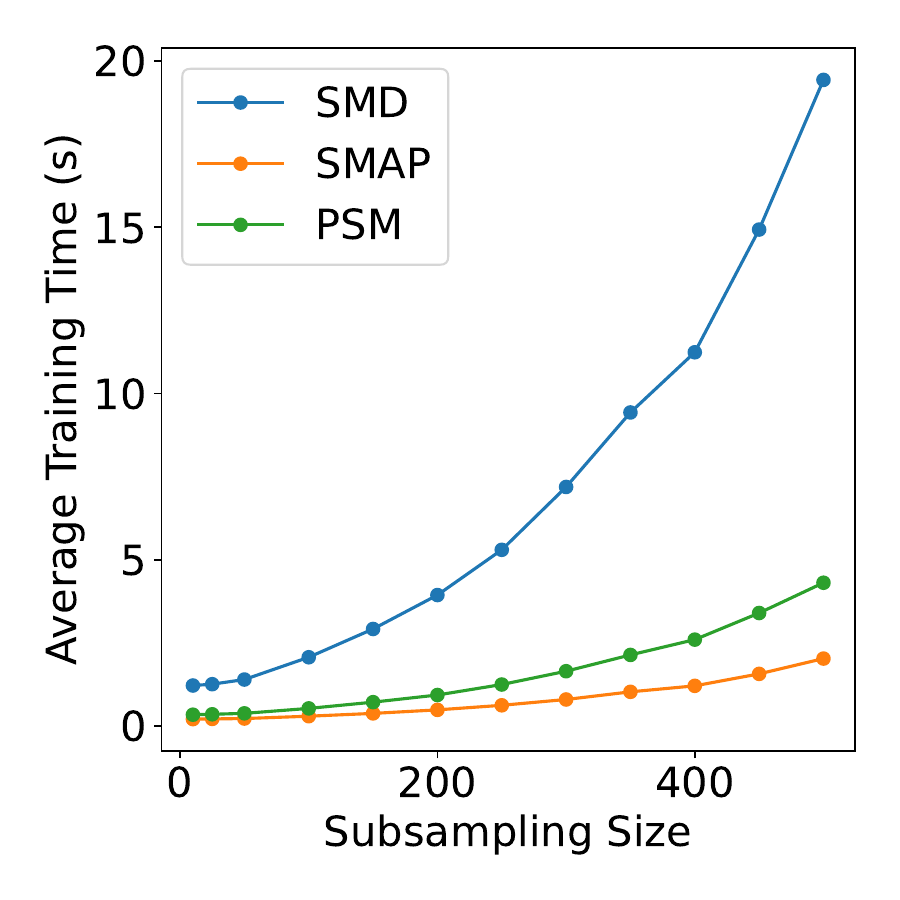}\label{fig:subsampling-time}}
	\caption{
		(a) Performance of IncFed MD-RS with different subsampling sizes $\tilde{N}_\mathrm{x}$ on three datasets, SMD, SMAP, and PSM.
		Solid curves show the average scores of PATE over the test time series for each dataset.
		The number of the reservoir nodes $N_\mathrm{x}$ is fixed at 500.
		(b) Average training time per time series for different subsampling sizes on three datasets, SMD, SMAP, and PSM.
	}
	\label{fig:subsampling-result}
\end{figure}

Fig. \subref*{fig:vary_client_size} demonstrates how the number of clients affects the model performance with four federated learning methods: FedAvg TranAD, IncFed ESN-SRE, FedAvg MD-RS, and IncFed MD-RS.
We calculate PATE scores with the varying number of clients.
The result shows that the two FedAvg-based methods, FedAvg TranAD and FedAvg MD-RS, achieve lower scores apparently as the number of clients increased while IncFed ESN-SRE and IncFed MD-RS are not affected by the number of clients.
It seems that the stability of the performance in IncFed MD-RS and IncFed ESN-SRE is due to the fact that both methods make it possible to construct mathematically equivalent models to the corresponding centralized models.

It is noteworthy that the performance of IncFed MD-RS is the same as that of centralized MD-RS in tables for every dataset, validating the equivalence of them.

This result suggests that the Incremental Federated Learning approach is well-suited for federated learning, when a large number of clients participate in the training.

\subsection{Subsampling}

Fig. \subref*{fig:subsampling-performance} shows the performance of IncFed MD-RS for different subsampling size $\tilde{N}_\mathrm{x}$ on the three datasets, SMD, SMAP, and PSM.
We measure PATE scores for every test time series and calculate the average score, represented by the circles on the solid curve.
The number of the whole reservoir nodes $N_\mathrm{x}$ is fixed at 500.

The result indicates that the performance of IncFed MD-RS is almost the same for the subsampling size from 100 to 500, showing that the performance of IncFed MD-RS is robust against subsampling $\tilde{N}_\mathrm{x}$ to some extent, which contributes to reducing the communication cost from the clients to the server.

Fig. \subref*{fig:subsampling-time} shows the average training time for each dataset.
The training time reduces as the subsampling size decreases across all datasets.
For SMD, which has the longest training data on average, subsampling is effective for enhancing computational efficiency.

\section{conclusions}
This research presented IncFed MD-RS, a novel federated learning scheme utilizing reservoir computing for time series anomaly detection.
In federated learning, clients often have limited computational resources and unreliable network connections, which pose significant challenges to the training process.
To address such issues, IncFed MD-RS provides an efficient method in terms of both computation and communication while protecting data privacy.
Moreover, the simplicity of reservoir state analyses enables an optimal aggregation of local models in the sense that the models produced through IncFed MD-RS are mathematically equivalent to the corresponding centralized models.
Our extensive experiments demonstrated that the proposed method outperforms other reservoir-based and deep learning-based approaches in terms of anomaly detection performance while being well-suited for federated learning, in which a large number of clients participate in the training process.

\section*{Code Availability}
The source code of IncFed MD-RS is publicly available at https://github.com/Key5n/IncFed-MDRS/.



\end{document}